# Decision Flexibility

**Tom Chávez and Ross Shachter**

**Department of Engineering-Economic Systems,
Stanford University**
chavez,shachter@bayes.stanford.edu

**Abstract:** The development of new methods and representations for temporal decision-making requires a principled basis for characterizing and measuring the **flexibility** of decision strategies in the face of uncertainty. Our goal in this paper is to provide a framework — not a theory — for observing how decision policies behave in the face of informational perturbations, to gain clues as to how they might behave in the face of unanticipated, possibly unarticulated uncertainties. To this end, we find it beneficial to distinguish between two types of uncertainty: "Small World" uncertainty and "Large World" uncertainty. The first type can be resolved by posing an unambiguous question to a "clairvoyant," and is anchored on some well-defined aspect of a decision frame. The second type is more troublesome, yet it is often of greater interest when we address the issue of flexibility; this type of uncertainty can be resolved only by consulting a "psychic." We next observe that one approach to flexibility frequently used in the economics literature is already implicitly accounted for in the Maximum Expected Utility (MEU) principle from decision theory. Though simple, the observation establishes the context for a more illuminating notion of flexibility, what we term **flexibility with respect to information revelation.** We show how to perform flexibility analysis of a static (i.e., single period) decision problem using a simple example, and we observe that the most flexible alternative thus identified is not necessarily the MEU alternative. We extend our analysis for a dynamic (i.e., multi-period) model, and we demonstrate how to calculate the value of flexibility for decision strategies that allow downstream revision of an upstream commitment decision.

## 1.0 Flexibility, Uncertainty, and Information

Researchers in decision-making under uncertainty have recognized the importance of developing models and methods that more adequately address the notion of **time**: How do decisions taken *now* shape or constrain decision opportunities confronted in the future? Whether we are building decision analyses to clarify human-oriented policy decisions or algorithms to approximate intelligent agency in a robot, the question poses a significant challenge for any methodology which claims to provide a complete, normative basis for action. Planning systems, for example, need rigorous methods for judging whether a particular plan is more "brittle" or less "flexible" than another, as well as sensible metrics for quantifying those features precisely. Decision consultants, meanwhile, increasingly focus on building strategies that create and sustain value for a decision maker or an organization over time, strategies that are somehow robust in the face of rapidly changing circumstances and unanticipated outcomes.

The reason we care about notions such as flexibility, brittleness, and robustness in decision-making is that the world is uncertain. If the world were perfectly deterministic — or even approximately so — we could simply order our actions to respond to future events using propositional logic. For the many problem areas where such reasoning falls short, we model uncertainty using the axiomatic framework of probability theory. Utility theory provides a normative basis for action using the probabilities assessed by a decision maker. Flexibility enters the discussion because we would like our actions to accommodate uncertain outcomes, and even better, to respond to uncertainties that we perhaps do not explicitly consider at the time we make a decision. As the mechanism through which we reduce uncertainty, **information** ought to figure prominently in any analysis of flexibility: given more information, we might do otherwise and thereby achieve higher value; the revelation of information downstream of a decision often creates or constrains opportunities to respond effectively to our upstream commitments.

Practicing decision analysts use a variety of sensitivity techniques to determine which sources of uncertainty weigh most heavily in the identification of optimal courses of action (see, for example, [Morgan and Henrion, 1990]). Deterministic perturbation, proximal analysis, and rank-order correlation are useful tools for measuring the relative importance of different uncertainties in a decision model. **Value of information** is the most powerful approach to sensitivity analysis because it measures not just whether uncertainty in an input variable could affect the output value, but rather whether reducing uncertainty in the input variable could change the recommended *decision*. Recent research provides efficient techniques for estimating information value in very large models [Chavez and Henrion, 1994].

All such sensitivity methods, however, occupy a separate phase of the decision analysis cycle or a special function of a decision support system, in both cases usually at what we might call the back end. They tell a decision maker which uncertainties matter, but they do not provide a necessary loop back to the front end: How should a decision maker use sensitivity measures to gain insight into the recommended action, or to craft better strategies? How should a decision maker use sensitivity results on *variables* to identify strengths and weaknesses in *decisions*? So far, sensitivity methods provide clues, but no compre-hensive basis, for measuring the relative robustness or flexibility of competing plans.

In this paper, we study the issue of flexibility using decision analysis. Our goal is to provide a framework — not a theory — for observing and measuring how decision policies respond to informational perturbations, to gain clues as to how they might respond to unanticipated, possibly unarticulated uncertainties. In Section 2, we distinguish between two central types of uncertainty, in an effort to delineate a more precise sense in which flexibility analysis is possible. In Section 3, we present and motivate a definition of flexibility developed in the economics literature, and we show how decision theory implicitly accommodates it. In simple terms, decision theory delivers one type of flexibility — what we call **flexibility with respect to values on outcomes ($F_{VO}$)**— "for free." Though simple, the analysis sets the stage for a more illuminating notion of flexibility which explicitly takes information and uncertainty into account: **flexibility with respect to information revelation ($F_{IR}$)**. In Section 4, we demonstrate how to measure this type of flexibility in a static (i.e., single period) decision model, using a canonical problem from decision analysis. We observe that the most flexible action thus identified is not necessarily the MEU (Maximum Expected Utility) alternative. In Section 5, we extend the analysis for a dynamic model which allows a downstream revision of an upstream commitment decision.

## 2.0  Distinctions about Uncertainty

We can think of flexibility generally as "the ability to adapt to changing circumstances" [Mandelbaum, 1978]. Changing circumstances are just the outcomes of random variables: for example, the interest rate suddenly dips, your car's electrical system suddenly blows a fuse, or your apartment is damaged in an earthquake. Adapting to changing circumstances requires that the actions you take now allow you to respond effectively to new discoveries or new situations in the future. As observed in [Ghemawat, 1991], "... a strategic option has flexibility value not because it is a sure thing but to the extent that it is an abundant store of potentially valuable revision possibilities." For example, if I am attempting to decide what car to buy, I might buy a Jeep instead of a Cadillac if I anticipate having to drive along treacherous mountain roads frequently in the future. If an agent is attempting to decide what resources to gather from its environment, it needs to deter-mine which resources will best ensure its long-term survival given reasonable expectations about future states of the world. We find it useful to fix ideas as follows:

"Flexibility is the ability to achieve greater value given the revelation of missing but knowable information downstream."

## 2.1    Small Worlds, Large Worlds, Clairvoyants, and Psychics

Missing but knowable information can assume several forms. In addressing issues of flexibility, we have found it useful to distinguish between two broad categories of information or uncertainty: **"Small World"** and **"Large World."**[1] We use the term "Small World" to bound the type of uncertainty to which it corresponds: Small World uncertainty can be resolved by observing an outcome on a variable which is *clearly and explicitly defined by a decision-making agent*. Decision analysts often use the concept of the **clairvoyant** to pose and resolve information questions. The clairvoyant is a thought experiment (similar to Maxwell's Demon, for example), a hypothetical person who can answer questions relevant to a decision problem. For example, if you are betting on coin flips, then the clairvoyant can tell you the outcome of the next flip. The clairvoyant cannot tell you whether you should consider betting on horses rather than coin flips. Nor can the clairvoyant offer information, e.g., he cannot suddenly tell you that the next coin flip will turn up something besides heads or tails. He can only *answer* unambiguous questions of fact.

Another type of missing but knowable information relates to those alternatives, uncertainties, or preferences that we have not explicitly defined or articulated. This type of information corresponds to Large World uncertainty. For example, knowing that the next coin flip will cause your opponent to suffer a heart attack at the gambling table certainly counts as missing, knowable information, but it probably is not anything we explicitly include in our betting analysis. In general, Large World uncertainties can be resolved only by posing open-ended questions to a **psychic**: for example, "Will anything strange happen when I next flip this coin?" Hypothetical answer: "Yes, your opponent will suffer a heart attack while it is in the air."

1. For treatment of a similar distinction, see [Laskey, 1992a,b].

Large World uncertainties are those uncertainties that we have not already specified, even though their outcomes could significantly affect the value we achieve.

Missing but knowable information provided by a clairvoyant corresponds exactly to the class of Small World uncertainties. Missing but knowable information provided by a psychic corresponds exactly to the class of Large World uncertainties.

## 2.2    Missionaries and cannibals revisited

The Large World/Small World distinction is key because it draws the line between what's possible and what's infeasible for flexibility analysis. We worry about flexibility in decision-making because we want our decision strategies to be responsive to a range of uncertain outcomes. But such strategies can be responsive only with respect to uncertainty that we deliberately, explicitly articulate.

It is perhaps useful to think of an analogue to the well-known AI Frame Problem in this context: Suppose you are charged with the task of transporting a group of missionaries safely between opposite sides of a river. Cannibals lurk in the bushes at both banks. A decision-analytic treatment of the problem would take account of the risks of encountering cannibals at different locations along the banks, given rustling movements in the bushes, say, and the disutility of getting shot by a cannibal's arrow. In implementing a decision strategy for transporting the missionaries, you might attempt to take action that flexibly accommodates all manner of bizarre risks. For example, you might worry about the possibility of an oar snapping in half or the possibility of a sudden, violent thunderstorm erupting while you are crossing the river.

Yet it is clearly impossible — and certainly impractical — to include all such remotely relevant, low-probability uncertainties in your analysis. Suppose for the moment that you construct a plan flexible enough to handle a snapping oar or a violent thunderstorm, and you begin to cross the river. In the middle of the river, your boat pops a leak and sinks; the missionaries drown and die, while the cannibals at the banks wail and wring their hands at their loss. Regardless of how flexible you *thought* your plan was, it certainly was not flexible enough to accommodate this bizarre circumstance.

It is important to observe that sometimes the unanticipated event can be beneficial. For example, the cannibals might be away for the wedding of their chieftain's daughter in a nearby village. Yet this is probably outside the realm of anything a reasonable person would include in a plan for transporting missionaries. Trying to accommodate all possible uncertainty — particularly Large World uncertainty — in flexibility analysis is infeasible, and probably incoherent in any case. The difficulty with this is that people often evaluate actions retroactively with respect to outcomes on Large World uncertainties. For example, if the boat pops a leak and sinks, then you are accused of implementing a "bad" plan to save the missionaries. If you are a decision analyst, then you respond that a leak in the boat was not even in the realm of discourse at the time you formulated the model; the event was vaguely relevant to the decision, but only marginally more relevant than a host of other variables which appear natural to eliminate from the analysis. The best we can do — and this is the thrust of the framework we propose in this paper— is to analyze flexibility with special attention to Small World uncertainty, in an effort to gain clues as to how strategies thus analyzed will behave in the face of Large World uncertainty.

## 3.0 Flexibility on Outcomes

We define a decision problem as a triple $[\Omega(X), \Omega(D), v(D,X)]$, where $X$ is a vector of state variables $[X_1,...,X_n]$, $\Omega(X)$ is the space of outcomes over $X$, and $D$ is a single decision with $m$ alternatives $[d_1,...,d_m]$. We will use capital letters to denote a variable or decision, and lower-case letters to denote a corresponding value or alternative, respectively: e.g., $D$ denotes the decision, and $d_1$ denotes a particular alternative for it; $X$ is a state variable, and $x$ denotes a particular value for $X$. We will also use $\Omega(X)$ to denote the set of possible values for $X$, where $X$ can be a state variable *or* a decision: e.g., $d \in \Omega(D)$ and $x \in \Omega(X)$. The value function $v(d,x)$ specifies the payoff/value/utility when action $d$ is taken and outcome $x \in \Omega(X)$ obtains.

$X$ may be defined probabilistically. If $X$ is a random variable, then $P\{X|\xi\}$ denotes a probability mass or probability density assignment on $X$, conditional on $\xi$, our prior state of knowledge. We will use $E[v(D,X)|\xi]$ to denote the expected value of $v(D,X)$. We will also use $E_X[v(D,X)|\xi]$ to denote the same measure, subscripting by $X$ to indicate that the expectation is taken with respect to $X$.

Stigler [1939] presents a useful and intuitive notion of flexibility, which has been studied and extended by a number of other researchers (see, e.g., Marschak and Nelson [1962], Jones and Ostroy [1984], Epstein [1980], and Merkhofer [1975]). Stigler characterizes the flexibility of two alternative plants using the second derivative of their total cost curves: a less flexible Plant A has a second derivative that is strictly greater than the second derivative corresponding to a more flexible Plant B, for all possible output values $X$. Figure 1 demonstrates the idea.

**FIGURE 1.**  Stigler's approach to flexibility.

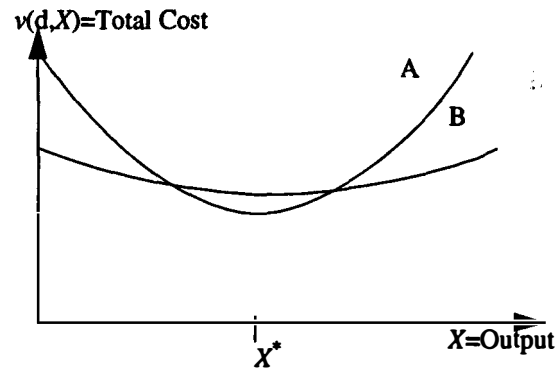

Both plants achieve minimum cost at the same level of output $X^*$, and Plant A actually beats Plant B for output $X^*$. Yet intuitively, it seems that we should prefer Plant B because of the flexibility it gives us over a broader range of possible output values for $X$.

Figure 2 shows the same kind of problem cast in decision terms. The decision depicted has three alternatives; the value curve for each is labeled by the decision alternative. In the sense introduced by Stigler, it appears that $d^+$ is most flexible in that it returns nearly constant value over the entire range of $X$.

**FIGURE 2.**   Stigler-type flexibility for a decision problem with three alternatives.

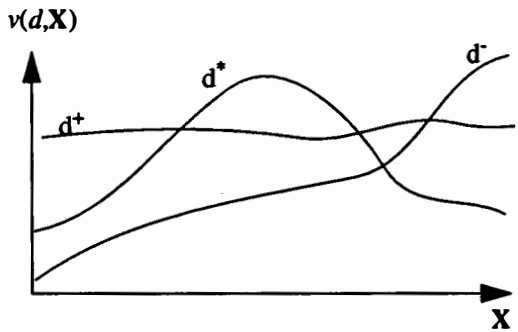

$v(d, \mathbf{X})$

We might say that $d^+$ is most flexible, or least brittle, in the sense that it minimizes the cumulative distance between itself and the upper boundary of the value curves for each of the available alternatives. Though $d^*$ and $d^-$ have value curves that peak more highly at separate points, $d^+$ does not suffer the severe dips in value that $d^*$ and $d^-$ show over broader ranges of values for $\mathbf{X}$. Because $\mathbf{X}$ is given probabilistically, we must weigh this cumulative distance by its probability of occurrence under $P\{\mathbf{X}|\xi\}$. In the spirit of Stigler, let us entertain the following definition.

**Definition 1.** The **brittleness** of action $d_i$ with respect to outcomes $\mathbf{X}$ is given by

$$\mathbf{E}_{\mathbf{X}}\left[\left(\begin{matrix} max \\ d \end{matrix}\right) v(d, \mathbf{X}) - v(d_i, \mathbf{X})\,\middle|\,\xi\right] \qquad \text{(EQ 1)}$$

The least brittle action is the one that minimizes the preceding quantity.

It is useful to observe the intended duality between the notions of flexibility and brittleness. Flexible strategies are less brittle because they do not "break" as easily in the face of unexpected outcomes. Brittle strategies are less flexible because they do not respond effectively to new discoveries or unanticipated outcomes.

Definition 1 seems to capture an intuitively appealing notion of flexibility following a standard approach from the literature. The following result shows that this notion of flexibility — which we will denote $\mathbf{F_{VO}}$, for flexibility with respect to outcomes — is already implicit in the Maximum Expected Utility (MEU) principle from decision theory.

**Theorem 1**: The optimal MEU action is the least brittle decision in the sense of Definition 1.

**Proof**: Suppose the least brittle action $d_b$ solves the equation

$$\text{(EQ 2)}$$

$$d_b = \begin{matrix} arg \ min \\ d_i \end{matrix} \left(\mathbf{E}_{\mathbf{X}}\left[\left(\begin{matrix} arg \ max \\ d \end{matrix}\right) v(d, \mathbf{X}) - v(d_i, \mathbf{X})\,\middle|\,\xi\right]\right)$$

We can expand the right-hand side as follows:

$$\text{(EQ 3)}$$

$$\begin{matrix} arg \ min \\ d_i \end{matrix} \mathbf{E}_{\mathbf{X}}\left[\left(\begin{matrix} arg \ max \\ d \end{matrix}\right) v(d, \mathbf{X})\,\middle|\,\xi\right] - \mathbf{E}_{\mathbf{X}}\left[v(d_i, \mathbf{X})\,\middle|\,\xi\right]$$

The first term inside the bracketed expression is a constant function, the upper envelope of the certainty equivalent curves for the different decision alternatives. Therefore minimizing the entire bracketed expression amounts to maximizing the second term inside the bracket — but this term is just the MEU decision, and the result follows. □

Theorem 1 shows that flexibility with respect to outcome is already embedded in the MEU principle from decision theory. In effect, any time one performs decision analysis, one gets $\mathbf{F_{VO}}$ for free. While the result might seem to trivialize our current notion of brittleness, it actually indicates where the useful work lies: to establish a clear notion of flexibility or brittleness which is tied to what we might call *variability over belief*, not simply variability over outcome.

## 4.0   Flexibility and Information Revelation: The Static Case

When we face the possibility of receiving missing but knowable information, we use Bayes' Rule to update our beliefs. It is precisely the prospect of receiving new information on a Small World uncertainty that leads us to revise — or wish that we could revise — an upstream commitment. If we do not have such information on-hand, we cannot update a probability assessment with which to make a more informed decision. For Large World uncertainty, the variable we would update has not even been introduced into the analysis.

If we are using parameterized probability distributions to represent our uncertainty, however, it is possible to ask how new information could affect our decisions *over the entire range of what we might discover*. Missing but knowable information allows us to update parameters on uncertainties within a decision model. Even if we do not have such information on-hand, we can nevertheless measure the extent to which decision strategies shift in response to perturbations in our beliefs about underlying uncertainties. Thus, flexibility analysis should focus on second-order uncertainty, and, in particular, sensitivity to second-order uncertainty (see, e.g., [Howard, 1988], [Pearl, 1988], [Heckerman and Jimison, 1987], [Chávez, 1995].)

For simplicity, assume a decision model with a single uncertainty $P\{X|\xi\}$ parameterized by a parameter $\pi$. For example, if $X$ is a binomial or geometric random variable, then $\pi$ could be the probability of success. Suppose that a decision maker assesses a value function $v(d,X)$ and a prior probability $\pi_o$ for $P\{X|\xi\}$. In Figure 3, we graph the certainty equivalent lines for three alternatives, i.e., $E[v(d,X)|\pi,\xi]$. Notice that, in contrast to Figure 2, we are in *expected-value* space (not value space), and that $E[v(d,X)|\pi,\xi]$ is always a straight line because expectation is a linear operator. Each certainty equivalent line is labeled by its corresponding decision alternative; $d^*$ is the MEU action.

**FIGURE 3.**    Certainty equivalent lines for a decision problem with three alternatives.

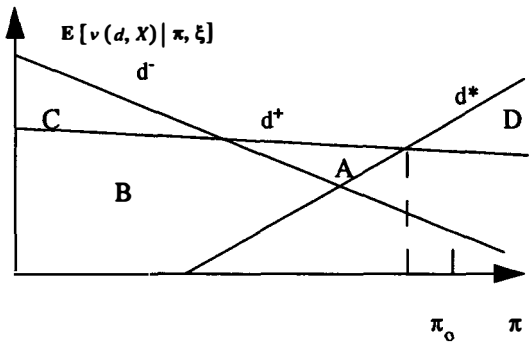

Given the decision maker's prior assessment $\pi_o$, $d^*$ is clearly the optimal alternative. If the decision maker receives new information which he uses to update $\pi$, then his optimal alternative quickly shifts; it seems logical to say that $d^*$ breaks too easily when new information becomes available.

Suppose $P\{\pi|\xi\}$ is the **maximum entropy** (i.e., uniform) distribution on the unit interval. The distribution reflects maximum uncertainty regarding what could be discovered about $\pi$, and thereby serves as the natural baseline of comparison for flexibility analysis. If we consider the upper envelope of the certainty equivalent curves as the best expected value the decision maker can achieve over the entire range of what he could discover about $\pi$, then $d^*$'s cumulative distance from that boundary equals (A+B+C), while $d^+$'s distance from the boundary is (D+C). For evocative purposes, we intend that (A+B+C)>>(D+C). These areas represent *the total expected value that the decision-maker could harness by doing otherwise given missing but knowable information*. In this sense, we assert that $d^+$ is less brittle than $d^*$ with respect to variability of belief. Or, equivalently, we say that $d^+$ is more flexible under information revelation than $d^*$.

**Definition 2:** Suppose $\Pi$ parameterizes $P\{X|\xi\}$, and that $P\{\Pi|\xi\}$ is the uniform distribution. The **brittleness of action $d_i$ with respect to variability of belief** $\Pi$ is given by

**(EQ 4)**

$$E_\Pi\left[\binom{max}{d} E_X[v(d,X)] - E_X(d_i,X)|\xi\right]$$

The least brittle action in this sense is the one that minimizes the preceding quantity.

## 4.1    An Example: The Party Problem

To see how this analysis works, consider the problem of determining where to have a party. The alternatives are "outdoors," "porch," and "indoors," and the uncertain decision variable is the weather, which can have two outcomes, either "rain" or "shine." Suppose that we represent our decision maker's utility via the following pay-off function (in dollars) with a set of decision alternatives $D$=[Por-

ch,Outdoors, Indoors] and two possible states of the world for weather, {rain, shine}.

**TABLE 1.**    Decision maker's payoff for Party Problem.

|          | Sun | Rain |
|----------|-----|------|
| Outdoors | 100 | 0    |
| Porch    | 90  | 20   |
| Indoors  | 40  | 50   |

Let $X$ denote the Bernoulli state variable on weather, and suppose that the decision maker assesses a 0.8 chance of sunshine on the day of the party. It is easy to verify that the optimal MEU decision $d^*$ is to have the party Outdoors, with expected value $80.

The decision maker is relatively confident about the probability of sunshine, and thus it is little surprise that the analysis suggest a party outdoors. He would like to understand his decision better, but he is reluctant to spend a lot of time analyzing it (buying groceries and setting up for the party are more urgent worries), and he is unwilling to spend too much time trying to examine all the possible evidence he could gather that would lead him to change his mind. He decides to perform a quick flexibility analysis, to answer the question, "How could missing but knowable information revealed *later* affect my ability to achieve value *now*?"

The breakpoint probabilities are .375 and .667, approximately, and it straightforward to verify that the certainty equivalent (CE) curves for the three alternatives are given as functions of $\pi$, the probability of sunshine, as follows:

Porch: $CE_P(\pi) = 20 + 70\pi$

Indoors: $CE_I(\pi) = 50 - 10\pi$

Outdoors: $CE_O(\pi) = 100\pi$

A sample calculation of the brittleness of the Indoors alternative with respect to variability of belief on $\pi$ proceeds as follows:

$$\int_{0.375}^{0.667} ((20 + 70\pi) - (50 - 10\pi)) \, d\pi + \int_{0.667}^{1.0} (100\pi - (50 - 10\pi)) \, d\pi$$

$$= [40\pi^2 - 30\pi] \Big|_{0.375}^{0.667} + [55\pi^2 - 50\pi] \Big|_{0.667}^{1.0}$$

$$= \$17.30$$

Similar analyses for Porch and Outdoors give brittleness measures of $7.29 and $12.74, respectively. Thus, our analysis indicates that Porch is the most flexible, or least brittle, alternative. Its brittleness measure, $7.29, represents the cumulative expected loss to the decision maker without information that could lead him to believe differently about the probability of sunshine.

### 4.2    Flexibility and Clairvoyance

Now consider the case where the decision maker is able to consult a clairvoyant free of charge. When the clairvoyant predicts "Sun," the decision maker moves the party outdoors to achieve value 100, while if the clairvoyant predicts "Rain," he moves the party indoors to achieve value 50. As a function of $\pi$, his certainty equivalent with free clairvoyance can thus be described as $100\pi + 50(1 - \pi) = 50 + 50\pi$. The reader can verify that this line connects the intercepts at $CE_O(\pi = 1)$ and $CE_I(\pi = 0)$.

A different but related measure of brittleness compares the cumulative difference in expected value that the decision maker achieves by sticking to a particular alternative and the expected value he achieves given free clairvoyance. The expected value given free clairvoyance is the gold standard for flexible action in a static model, because it represents the value the decision maker achieves by gathering *all* missing but knowable information free of charge before taking action.

**Definition 3**: Suppose that $\Pi$ parameterizes $P\{X | \xi\}$, and that $P\{\Pi | \xi\}$ is the uniform distribution. The **brittleness of action $d_i$ with respect to variability of belief $\Pi$, given clairvoyance on X,** is

$$E_{\Pi, X}\left[\binom{max}{d}(v(d, X) | \xi)\right] - E_{\Pi, X}[v(d_i, X) | \Pi, \xi] \quad \text{(EQ 5)}$$

The least brittle action in this sense is the one that minimizes the preceding quantity.

The measure of brittleness in this sense preserves exactly the ordering on alternatives implied by Definition 2, and is always greater than or equal to the brittleness measure given in Definition 2. In general, it seems to be slightly easier to calculate. An example for Porch is given below:

$$\int_0^1 ((50+50\pi)-(20+70\pi))\,d\pi$$

$$= [30\pi - 10\pi^2]\,|_0^1$$

$$= \$20$$

Measures for Indoors and Outdoors are $30 and $25, respectively.

## 5.0  Flexibility and Information Revelation: The Dynamic Model

Recall that our primary concern in studying flexibility is to take action now which allows us to respond effectively to changing circumstances, and, in particular, to make optimal use of information revealed in the future. Consider the simple two-period decision model depicted in Figure 4.

**FIGURE 4.**  The general influence diagram for flexibility in a dynamic model.

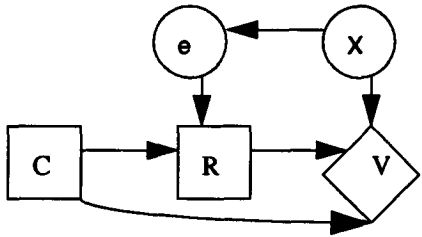

This simple decision scenario is characterized by a commitment decision $C$, a reaction or revision decision $R$ which temporally follows $C$, uncertainty $X$, information/evidence $e$ relevant to $X$ which arrives before $R$ is taken, and a value function $v(c,r,X)$. The arc from $C$ to $R$ denotes the fact that we remember the decision we took at $C$ when we are downstream at decision $R$.

Suppose now that $C$ has one of three possible alternatives $\{d^+,d^-,d^*\}$, and that only $d^+$ allows revision at $R$. In other words, $d^-$ and $d^*$ are strong commitments in the sense that they do not allow revision, while $d^+$ allows the possibility of revision to $d^-$ or $d^*$ at $R$. It is important to observe that $d^+$ is not necessarily more flexible than the other alternatives; the possibility to revise yields greater flexibility only if it allows us to achieve greater value downstream than we could have achieved otherwise. The option to revise is typically never free, first because the information gathered at $e$ will usually cost money or resources, and second because most real world revision decisions incur what we might call "switching costs." Figures 5 and 6 illustrate two possible sets of certainty equivalent curves for the decision model depicted in Figure 4. We assume that $X$ is binomial, that the decision maker's prior on $X$ is $\pi_o$, and that after evidence $e$, the decision maker's posterior on $X$ is either $\hat{\pi}$ or $\bar{\pi}$. The cost of revision plus the cost of information $e$ is $w$.

**FIGURE 5.**  Scenario 1: Revision at $R$ does not carry superior flexibility value.

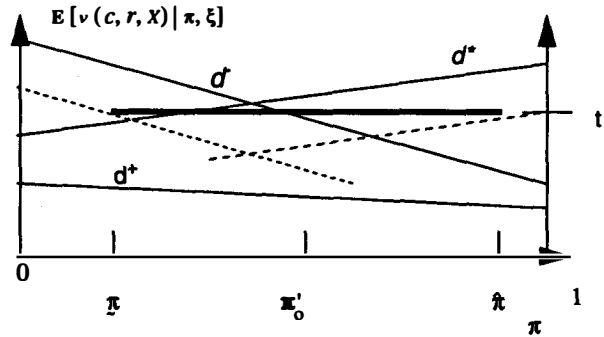

**FIGURE 6.**  Scenario 2: Revision at R carries superior flexibility value.

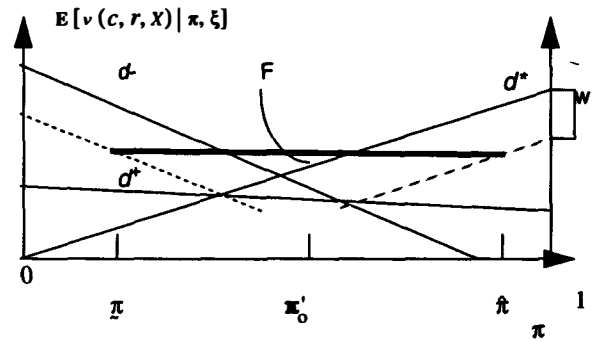

In both figures, dotted lines represent the certain equivalents of the decision alternatives $d^*$ and $d^-$ minus the cost $w$. In both, we represent the certainty equivalent line with information $e$ by a heavy line; similar to the case with clairvoyance presented in Section 4.2, it is the line connecting the two outer intercepts of expected value achieved with information. For example, if we perform $d^+$ at $C$ and then update our belief about $\pi$ to equal $\hat{\pi}$ after observing $e$, then we revise our decision to $d^*$ and achieve expected value at the level $t$ shown in Figure 5. Traveling along this heavy line and taking its value at $\pi = \pi'$, our prior assessment, gives the value with flexibility of alternative $d^+$. Since this point is below the expected value achieved with $d^*$, $d^+$ does not yield superior flexibility value in scenario 1.

In scenario 2, however, an analogous argument shows that $d^+$ yields superior flexibility value. After information $e$, we are able to achieve higher expected value in revising our decision than if we stick to a commitment $d^*$ or $d^-$. The value of the flexibility in $d^-$ is just the difference indicated by $F$ in Figure 6. We formalize this idea in the following definition.

**Definition 4**: We are given a decision model with temporally ordered decision $C$ and $R$, information $e$ revealed before $R$ and after $C$, and value function $v(c,r,X)$. Information $e$ is relevant to $X$. Then the **flexibility value of a commitment $c$ with respect to the revelation of information $e$, denoted $F_e(c)$**, is given by

$$\text{(EQ 6)}$$

$$\max_c \; E_e\!\left[\max_r \; E_X\left[v(c,r,X)\,|\,e,\xi\right]\right] - \max_{c,r} \; E_X\left[v(c,r,X)\,|\,\xi\right]$$

where the outside expectation of the first term is taken with respect to $e$ and the inside expectation is with respect to $X$. The commitment which maximizes this quantity for values strictly greater than zero is the most flexible commitment.

### 5.1  Example

Let us return to the Party Problem. Suppose now that the decision maker has the opportunity to commit to having the party outdoors or indoors; or he can accept a "soft" commitment to having the party on the porch, with the possibility of revising the porch decision later. If he chooses the porch now, he will have the opportunity to receive a meteorologist's report the day of the party. Suppose that the meteorologist's report costs only $1, and he further estimates a switching cost of $5 to move plates and tables from the porch to his living room or from the porch to his backyard. Hence the total cost to revise his decision is $6.

Our decision maker is a bit less confident of the probability of sunshine than before. He estimates the probability of rain at .7; the optimal action is still to have the party outdoors with expected payoff $70, if he chooses not to accept the Porch option. Suppose that the metereologist's one-day forecasts are highly reliable, with accuracy probability .9 (i.e., if he says "Rain," then it rains with probability .9, and if he says "Sun," then it is sunny with probability .9).

Given these assessments, the decision maker's posterior on rain given a "Rain" report from the meteorologist is .79, and his posterior on sun given a "Sun" report is .95. Thus, if the meteorologist says "Sun," the decision-maker moves outdoors with expected payoff $95-$6=$89, and if he says "Rain" he goes inside with expected payoff $47.9-$6=$41.9. Taking expectation over the preposterior on evidence — in this example, the reader can verify that the preposterior probability that the meteorologist reports "sun" is .66 — we find that the expected payoff to the decision maker with the Porch option is $72.9. This number is the first term in Definition 4. The flexibility value is the difference between this $72.9 and $70, or roughly three dollars.

Thus, in this example, the flexible alternative is the superior alternative, with an expected value difference to the decision maker of about $3.

## 6.0  Conclusions and Future Directions

The definition of flexibility as the ability to achieve value given the future revelation of missing but knowable information allows us to examine decision strategies in a new light. We have shown that MEU fully captures one type of flexibility, what we have termed flexibility with respect to values on outcomes. We have also identified a distinct, possibly useful, notion of flexibility, what we call flexibility with respect to information revelation. By anchoring our analysis on the entire *range* of beliefs that we might hold about our uncertainty, it is possible to measure the relative brittleness or flexibility of alternate strategies as the change in expected value to the decision maker, given information that could lead him to act differently. Static and dynamic analyses of decision scenarios allow us place precise flexibility values on different strategies, in a manner which confirms intuition for the simple problem considered here. Further work will focus on applying the framework developed here to larger models with many decisions and on exploring more of the framework's theoretical implications.

*Acknowledgments*: The second author would like to acknowledge John Mark Agosta, and both authors gratefully acknowledge Marvin Mandelbaum, for their help in stimulating and clarifying many of the ideas presented here. Kevin Soo Hoo and Wolfgang Spinnler made several useful comments on earlier drafts of this paper.